\documentclass[times, review, 10pt]{elsarticle}
\usepackage{amssymb}
\usepackage{amsmath}
\usepackage{longtable}% for long tables
\usepackage{mdframed} % 创建带边框的文本块
\usepackage{xcolor} % 颜色文字
\usepackage{makecell} % 表格单元格
\usepackage{multirow} % 表格跨行合并
\usepackage{fancyhdr}
\usepackage{booktabs}
\usepackage{lineno}
\usepackage{subfigure}
\journal{Nuclear Physics B}

\begin{document}

\begin{frontmatter}



\title{Physical Plausibility Reasoning via HCM-GRPO: Empowering Compact Model for Superior Performance}


% \author{Zheng Sun$^a$, Zhiyuan Hu$^b$, Yi Wei$^a$, Long Yu$^a$} %% Author name

% \affiliation{organization={Alibaba Group},%Department and Organization
%             %addressline={}, 
%             city={Beijing},
%             postcode={100102}, 
%             %state={},
%             country={China}}
% \affiliation{organization={Tsinghua University},%Department and Organization
%             %addressline={}, 
%             city={Beijing},
%             postcode={100084}, 
%             %state={},
%             country={China}}   

\author{
    % 第一行：作者姓名和角标
    Zhiyuan Hu$^{a,*,1}$, Zheng Sun$^{b,*,\dagger}$, Yi Wei$^b$, Long Yu$^{b,\dagger}$ \\[1ex] 
    % 第三行：地址a，用 \textit 斜体
    \textit{$^a$Tsinghua University, Beijing, 100084, China}\\
    % 第二行：地址b，用 \textit 斜体
    \textit{$^b$Alibaba Health Information Technology Limited, Beijing, 100102, China} \\
    %\textit{\{banqun.sz, huayuan.hzy, wy271630, yl185725\}@alibaba-inc.com}
}

% \author{
%     % 第一行：作者姓名和角标
%     Zheng Sun$^{a,*,\dagger}$, Zhiyuan Hu$^{b,*,1}$, Yi Wei$^a$, Long Yu$^{a,\dagger}$ \\[1ex] 
%     % 第二行：地址b，用 \textit 斜体
%     \textit{$^a$Alibaba Health Information Technology Limited, Beijing, 100102, China}\\ 
%     % 第三行：地址a，用 \textit 斜体
%     \textit{$^b$Tsinghua University, Beijing, 100084, China}\\
%     %\textit{\{banqun.sz, huayuan.hzy, wy271630, yl185725\}@alibaba-inc.com}
% }

%% Abstract
\begin{abstract}
The performance of image generation has been significantly improved in recent years. However, the study of image screening is rare, and its performance with Multimodal Large Language Models (MLLMs) is unsatisfactory due to the lack of data and the weak physical plausibility reasoning ability in MLLMs. In this work, we propose a complete solution to address these problems in terms of data and methodology. For data, we collect a comprehensive image screening dataset with over 128k samples, comprising about 640k images. Each sample consists of an original image and four generated images. The dataset evaluates the physical plausibility reasoning ability under four aspects: appearance deformation, physical shadow, placement layout, and extension rationality. Regarding data annotation, we investigate multiple approaches, including purely manual, fully automated, and answer-driven annotations, to acquire high-quality chains of thought (CoT) data in the most cost-effective manner. Methodologically, we introduce a Hard Cases Mining (HCM) strategy with a Dynamic Proportional Accuracy (DPA) reward into the Group Relative Policy Optimization (GRPO) framework, called HCM-GRPO. This enhanced method demonstrates superior physical plausibility reasoning capabilities compared to the original GRPO. Our experimental results reveal that even state-of-the-art closed-source MLLMs, such as GPT5.2 and Gemini3-Pro, exhibit unsatisfactory performance in physical plausibility reasoning. In contrast, by leveraging the HCM-GRPO, we are able to surpass the scores of both large-scale open-source and leading closed-source models with a much smaller model.

% Definitions for the numbered footnotes ('1' and '2')
\begingroup
\renewcommand\thefootnote{*}
\footnotetext{Equal contribution.}
\endgroup

% Definition for the '*' (Corresponding author)
\begingroup
\renewcommand\thefootnote{$\dagger$}
\footnotetext{Corresponding author.}
\endgroup

\begingroup
\renewcommand\thefootnote{1}
\footnotetext{Work done when Zhiyuan Hu was an intern at Alibaba Health Information Technology Limited, Beijing, China.}
\endgroup
% --- End of Footnote Definitions ---
\end{abstract}


% \begin{highlights}
% \item We establish a benchmark for screening AI-generated images focused on physical space.
% \item We propose HCM-GRPO to unlock the significant potential of compact models.
% \item The experiments demonstrate that our method surpasses that of large leading models.
% \end{highlights}

%% Keywords

\end{frontmatter}

\section{Introduction}
In recent years, there have been extensive researches on Multimodal Large Language Models (MLLMs), covering foundational model development~\cite{chen2024internvl, qwen2.5-VL}, evaluation dataset construction~\cite{yang2025cc, cheng2025comt}, reinforcement learning (RL) applications~\cite{liu2025visual}, and even areas related to AI-Generated Content (AIGC)~\cite{sun2024generative, tian2024visual, huang2025wegen}. At the same time, thanks to the development of diffusion models~\cite{ho2020denoising, rombach2022high, peebles2023scalable, ye2023ip} and unified MLLMs~\cite{huang2025wegen, wang2026multimodal}, the performance of image generation has also been greatly improved. The diffusion process of images involves a certain degree of randomness, so it often requires specific conditions to provide directed control~\cite{zhang2023adding}. However, even with constraints on the generation process, the model may still produce some unpredictable results. Therefore, it is highly necessary to conduct a screening process for the generated images. MLLMs are capable of processing information from different modalities, such as text, image and video, simultaneously, and providing responses based on a comprehensive understanding. Against this background, this paper focuses on exploring the physical plausibility reasoning ability of MLLMs for image screening.\\

\begin{figure}[!t]
\centering

\subfigure[Four evaluation dimensions of physical plausibility.]{
    % 直接在这里设置图片宽度
    \includegraphics[width=0.9\linewidth]{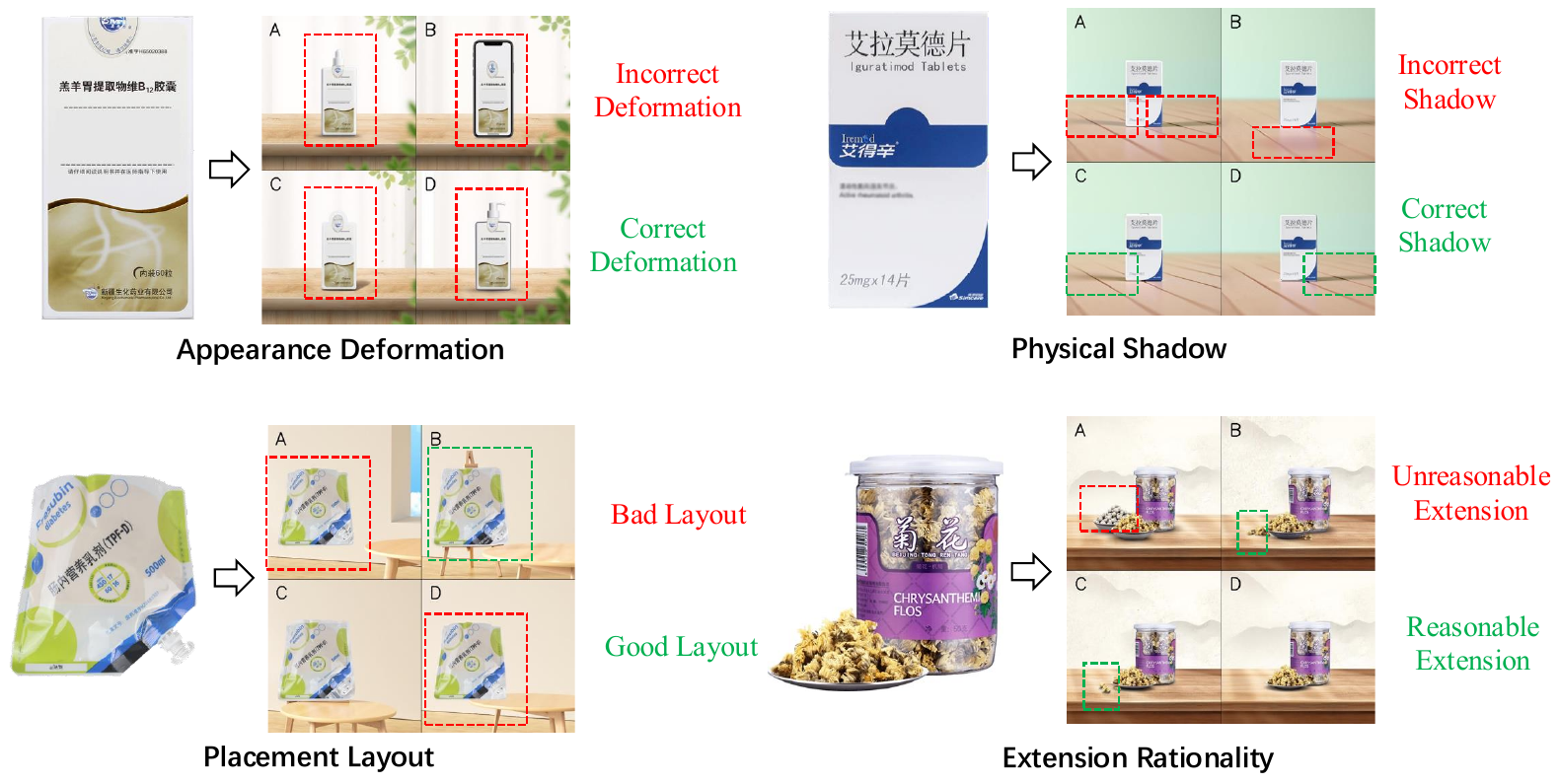}
}%
\quad
\subfigure[Quantitative comparison results.]{
    % 直接在这里设置图片宽度
    \includegraphics[width=0.9\linewidth]{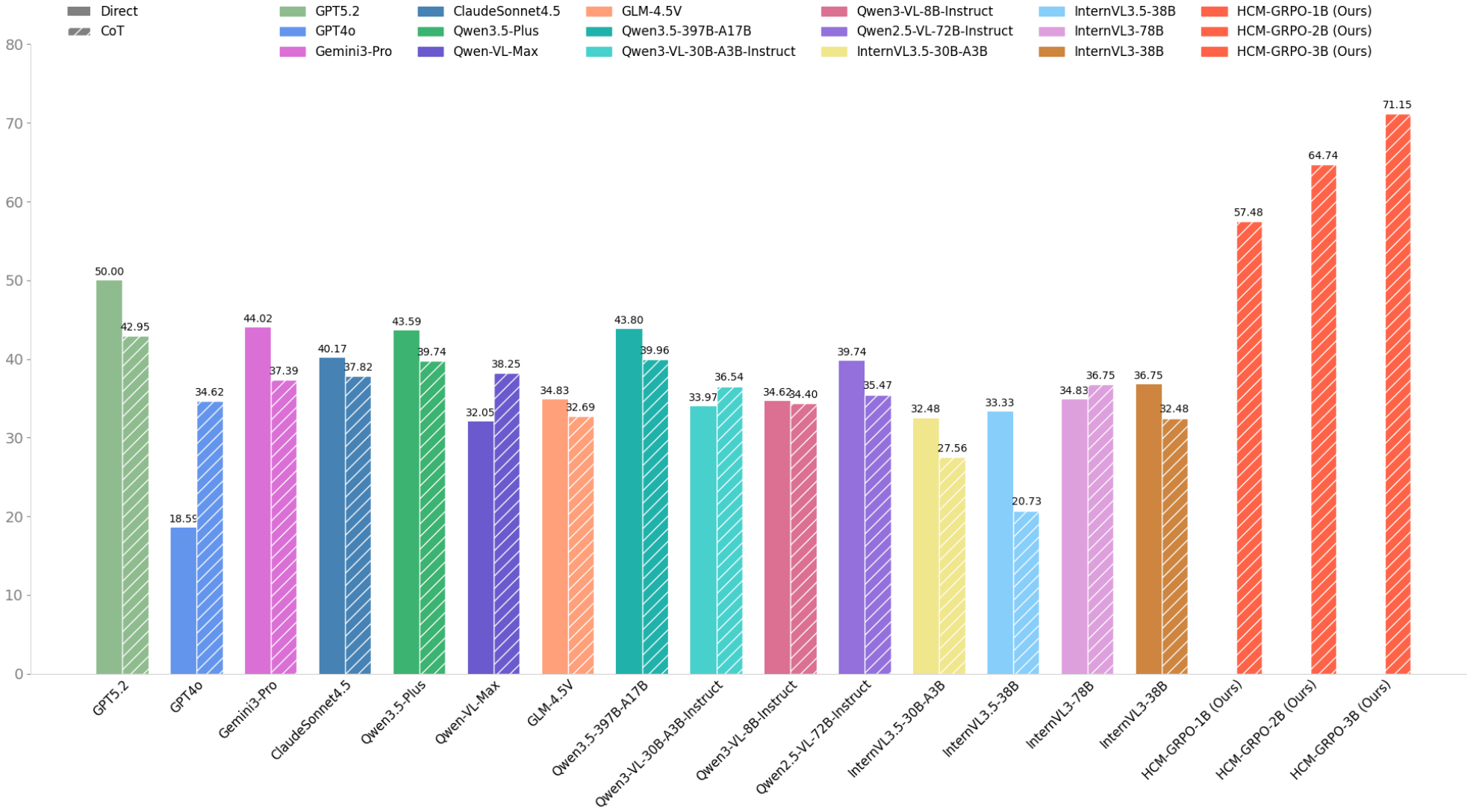}
}%

\centering
\caption{Overview of the physical plausibility dataset and quantitative comparison results. (a) We summarize four evaluation dimensions of physical plausibility from the dataset. (b) Extensive quantitative comparison results demonstrate the superiority of our HCM-GRPO method in the image screening task.} \label{Overview}
\end{figure}

Physical plausibility reasoning with MLLMs is hindered by the scarcity of specialized datasets and the suboptimal reasoning abilities of existing models. In response, advancements have been achieved through the development of a novel dataset and an advanced approach. The collected dataset in this paper consists of over 128k samples, providing indispensable knowledge for model training. In prior studies~\cite{tan2025ominicontrol, li2023theme, bin2024gallerygpt}, the definition of image aesthetics has been highly subjective. For instance, Subjects200K~\cite{tan2025ominicontrol} employs GPT4o to evaluate image quality, while TAVAR~\cite{li2023theme} leverages image themes to enhance the explainability of image aesthetics. Instead, our dataset places significant emphasis on the physical space transformations of AI-generated images. These transformations are straightforward to evaluate objectively, avoiding any reliance on subjective artistic criteria. The generated images consist of the foreground medicines, backgrounds, and layout settings. All the medicines are derived from the real world. We randomly select a background image from the background image set and randomly assign a layout (either top-bottom or left-right) for the original medicine image. Then, we use a segmentation model~\cite{zheng2024bilateral} on the original image to obtain the mask of the foreground. Based on the assigned layout region, we determine the position of the foreground target. Finally, we use an image generation model~\cite{flux2024} to render the areas outside the foreground medicine, referencing the selected background image. In the generated images, we observe a significant amount of unintended content due to randomness. As shown in Figure~\ref{Overview} (a), we categorize these issues into four types: appearance deformation, physical shadow, placement layout, and extension rationality. We summarize the ability to identify these four types of issues as physical plausibility reasoning capability. After collecting the images, we assign the four generated images the labels A, B, C, and D, and manually annotate the correct images in the form of multi-answer labels in the training and testing datasets, such as "AC", "BCD", or "N". We test various state-of-the-art closed-source and open-source MLLMs for the identification of problematic images, with the final results presented in Figure~\ref{Overview} (b). From the results, we conclude that directly using existing models yields unsatisfactory performance. Based on this, we further propose a two-stage approach to enhance the physical plausibility reasoning capability of small-sized MLLMs. Specifically, in the first stage, we investigate multiple approaches to acquire chains of thought (CoT) data to perform supervised fine-tuning (SFT) on the base model to adapt it to the specific response format. Purely manual annotation produces the highest quality CoT data but incurs high associated costs and a slow pace, making it difficult to scale. Fully automated annotation is extremely low-cost and highly scalable. However, it sacrifices quality, often generating noisy data that contains factual errors or hallucinations. Answer-driven annotation serves as a strategic trade-off. It leverages the final answer from humans and the thought process from existing models to accelerate the process and reduce costs. It balances the efficiency of automation with the reliability of human expertise. In the second stage, we propose Hard Cases Mining (HCM) in Group Relative Policy Optimization (GRPO)~\cite{guo2025deepseek} with Dynamic Proportional Accuracy (DPA) reward, called HCM-GRPO, to stimulate the model's reasoning ability. The HCM-GRPO first utilizes an initial model to partition the dataset into easy and hard cases, then it strategically focuses on these hard cases during the latter stage of training through hard cases mining. Concurrently, the integrated DPA reward provides a significantly denser reward signal to guide the optimization process more effectively. Ultimately, our approach achieves a score of 71.15 on the evaluation dataset with Qwen2.5-VL-3B~\cite{qwen2.5-VL}, surpassing large-sized open-source and leading closed-source models. Moreover, our experimental results on public datasets validate the applicability of HCM-GRPO to real-world and multi-image understanding tasks. The main contributions of this paper can be summarized as follows:
\begin{itemize}
% \item We develop a pipeline for image fusion and generation, curating an image screening dataset with more than 640k images, which is specifically designed for training and testing MLLMs, as well as for the proposed method for physical plausibility reasoning. Our dataset is built upon authentic, real-world images of medicinal products.
\item Few existing benchmarks offer multi-answer labels for physical plausibility visual reasoning. We address this by constructing a large-scale dataset for physical plausibility reasoning, built with an image fusion and generation pipeline and annotated with multiple correct answers.
\item We present a two-stage approach, which includes a scalable methodology for acquiring high-quality CoT data and an enhanced, more effective reinforcement learning method, HCM-GRPO. This methodology achieves a significant enhancement compared to the performance of the base model. The small models surpass both large-sized open-source and closed-source models on the scoring metric.
\item We conduct extensive experiments on established and other public benchmarks. The experimental results reveal that current MLLMs still have significant deficiencies in understanding physical plausibility, and our HCM-GRPO can effectively enhance the performance of the base model. It is hoped that this work will provide a valuable reference for research in the MLLMs and AIGC communities. 
\end{itemize}
%\acks{Acknowledgements should go at the end, before appendices and references. You can uncomment this for the camera-ready version on paper acceptance.}

\section{Related Work}
\subsection{Multimodal Large Language Models}
MLLMs have demonstrated impressive capabilities across various tasks and applications~\cite{chen2024internvl, liu2025visual, hong2025glm}. Particularly, an increasing number of studies are exploring the scope and boundary of MLLMs' capabilities. For instance, CC-OCR~\cite{yang2025cc} investigates the performance of MLLMs on end-to-end OCR tasks. Naturalbench~\cite{li2024naturalbench} examines their ability to understand when faced with the same question under different images. IntentMLM~\cite{wang2025unlocking} explores MLLMs for intent perception by leveraging their rich cross-modal representations. CS-VSL~\cite{zhang2025cross} explores cross-scene visual context parsing by leveraging MLLMs to capture relationships between objects across multiple images. SWG-Fusion~\cite{wang2026swg} leverages MLLMs to extract semantic weather cues from visual inputs and provide adaptive guidance for multimodal fusion in BEV object detection under harsh weather conditions. FER~\cite{li2026event} leverages MLLMs to perform facial expression recognition by integrating visual event streams and textual priors to model facial action unit correlations. CoMT~\cite{cheng2025comt} explores their visualization capabilities during the reasoning process. Inspired by the frequent occurrence of unreasonable content in the field of image generation, we are very curious about whether MLLMs can identify physically plausible generated contents. In this work, we construct a pipeline for the image generation dataset and evaluate the performance of various MLLMs based on this dataset.

\subsection{Reinforcement Learning}
Recently, reinforcement learning techniques have been extensively applied to enhance the reasoning capabilities of Large Language Models (LLMs), enabling them to effectively solve complex problems~\cite{guo2025deepseek, yu2026dapo}. DeepSeek-R1~\cite{guo2025deepseek} is a milestone work in the application of reinforcement learning to the domain of LLMs. It includes the full version of DeepSeek-R1 and DeepSeek-R1-Zero, which are obtained using Group Relative Policy Optimization (GRPO). Based on GRPO, DAPO~\cite{yu2026dapo} introduces several key techniques to make RL shine in the long-CoT RL scenario with Qwen2.5-32B, outperforming DeepSeek-R1-Zero-Qwen-32B while using only half the training steps in AIME 2024. As for the LLM agent, MCMIX~\cite{liu2026value} leverages deep reinforcement learning to address the non-monotonic value decomposition problem by robustly learning from multiple high-quality joint actions in cooperative multi-agent environments. As for MLLMs, RL is often applied to specific tasks, such as object detection~\cite{liu2025visual}, training reward models~\cite{wangunified}, and enhancing reasoning capabilities~\cite{Zhang_2025_ICCV}, which often require designing task-specific rewards. In this paper, we apply the reinforcement learning method GRPO to a new task: screening generated images with MLLMs. The proposed HCM-GRPO method, even when applied to small-sized models, outperforms both large-sized open-source and closed-source models.

% \begin{figure}[!t]
% \centering

% \subfigure[Construction pipeline of image screening dataset.]{
%     \includegraphics[width=0.6\linewidth]{fig2_1.pdf}

% }
% \quad
% \subfigure[Distribution of different evaluation dimensions.]{
%     \includegraphics[width=0.3\linewidth]{fig2_2.pdf}
% }%

% \centering
% \caption{Overview of dataset construction pipeline and proportion distribution of different evaluation dimensions in the dataset.} \label{Dataset}
% \end{figure}

\begin{figure}[!t]
\centering
\includegraphics[width=0.98\textwidth]{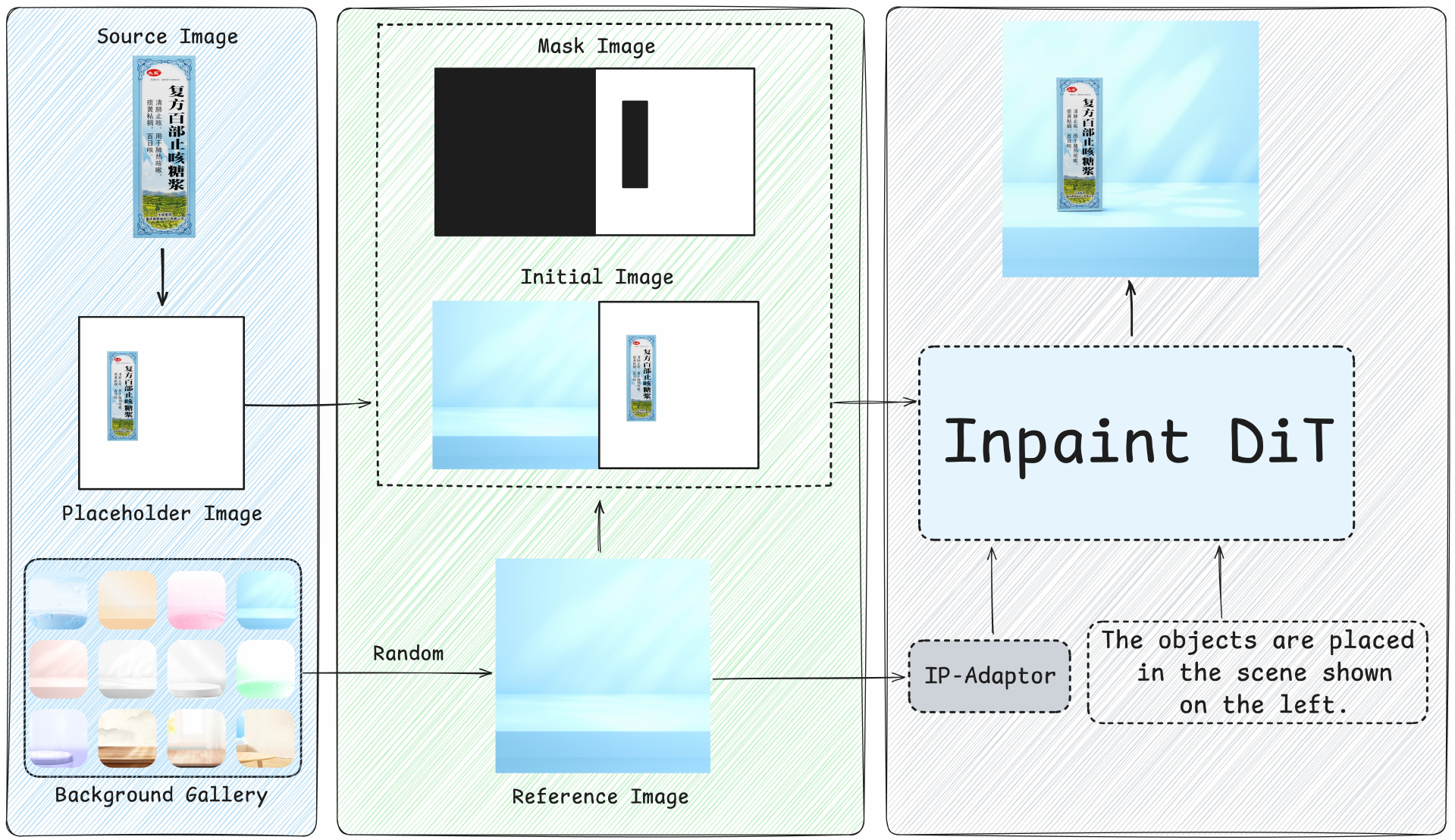}
\caption{Overview of dataset construction pipeline.} \label{Dataset}
\end{figure}

\section{Image Screening Dataset}
\subsection{Image Fusion and Generation}
In this section, we mainly introduce the data construction pipeline and provide detailed information about the dataset. The construction process can be divided into three stages, as shown in Figure~\ref{Dataset}. The first stage is the data preparation phase. We first obtain source images of various medicines and use an open-source image segmentation model~\cite{zheng2024bilateral} to extract the foreground regions of the medicines. These regions are then placed onto a preset area of a white background image to create the placeholder image. Meanwhile, we randomly select a background image from the background gallery as the reference image. The second stage is the data processing phase. We combine the placeholder image and the reference image to create the initial image. Additionally, we generate a mask image based on the placeholder image. The mask image consists of two parts: the left half is entirely black, while the right half is white, except for the target region. Notably, we attempt to use only the right half of the mask image and the initial image as input, however, the final rendering results are unsatisfactory. Therefore, both the initial image and the mask image include both the left and right halves. The third stage is the generation phase. Here, we utilize FLUX.1-Fill-dev and FLUX.1-Redux-dev~\cite{flux2024} as Inpaint DiT~\cite{peebles2023scalable} and IP-Adapter~\cite{ye2023ip}, respectively. The reference image is fed into the IP-Adapter to serve as a controller, while the mask image and the initial image are simultaneously input into the Inpaint DiT for repainting. In summary, the process involves redrawing the initial image with the white regions in the mask image by referencing the reference image.\\
Based on the pipeline described above, we collect over 640k images, which are ultimately grouped into more than 128k samples. The dataset contains over 56,500 types of medicines, more than 20 background images and 4 evaluation dimensions. As shown in Figure~\ref{Overview} (a), these four evaluation dimensions are as follows: (1) appearance deformation, which refers to differences in appearance between the generated image and the original medicine; (2) physical shadow, which indicates noticeable errors in lighting and shadow, such as inconsistent shadow directions among objects; (3) placement layout, which refers to unrealistic scenarios, such as floating objects in the image; (4) extension rationality, which involves the principle of generating logically consistent associations while strictly prohibiting fabrications that lack any grounding in the original medicine image. We summarize the ability to identify these four types of issues as physical plausibility reasoning capability. The most common issue during the image generation process is appearance deformation, and only a small portion (8.5\%) of the images are normal. Excluding the normal part, we focus on evaluating the ability of MLLMs to recognize issues across the remaining four dimensions. More visualization examples can be found in Figure~\ref{Demonstration1}.

\begin{table}[!t]
\centering
\renewcommand{\arraystretch}{0.85} 
% \small
\caption{An overview of the different dataset splits, detailing their characteristics such as size, label accuracy, CoT data, and supervision type.}
\label{tab:dataset_overview}
\resizebox{\linewidth}{!}{
    \begin{tabular}{ccccc}
    \toprule
    % 使用 \makecell 来允许表头换行
    \textbf{Dataset Split} & \textbf{Training} & \textbf{Testing} & \textbf{Pseudo-Label} & \textbf{Exploration} \\
    \midrule
    Size & 1,044 & 468 & 10,724 & 115,809 \\
    Multi-answer Label & \checkmark & \checkmark & $\times$ & $\times$ \\
    CoT Data   & \checkmark & $\times$ & \checkmark & $\times$ \\
    Supervision Type    & Fully + Answer-driven & Fully & Weakly & Unsupervised \\
    \bottomrule
    \end{tabular}
}
\end{table}

\begin{figure}[h]
\centering
\includegraphics[width=0.95\textwidth]{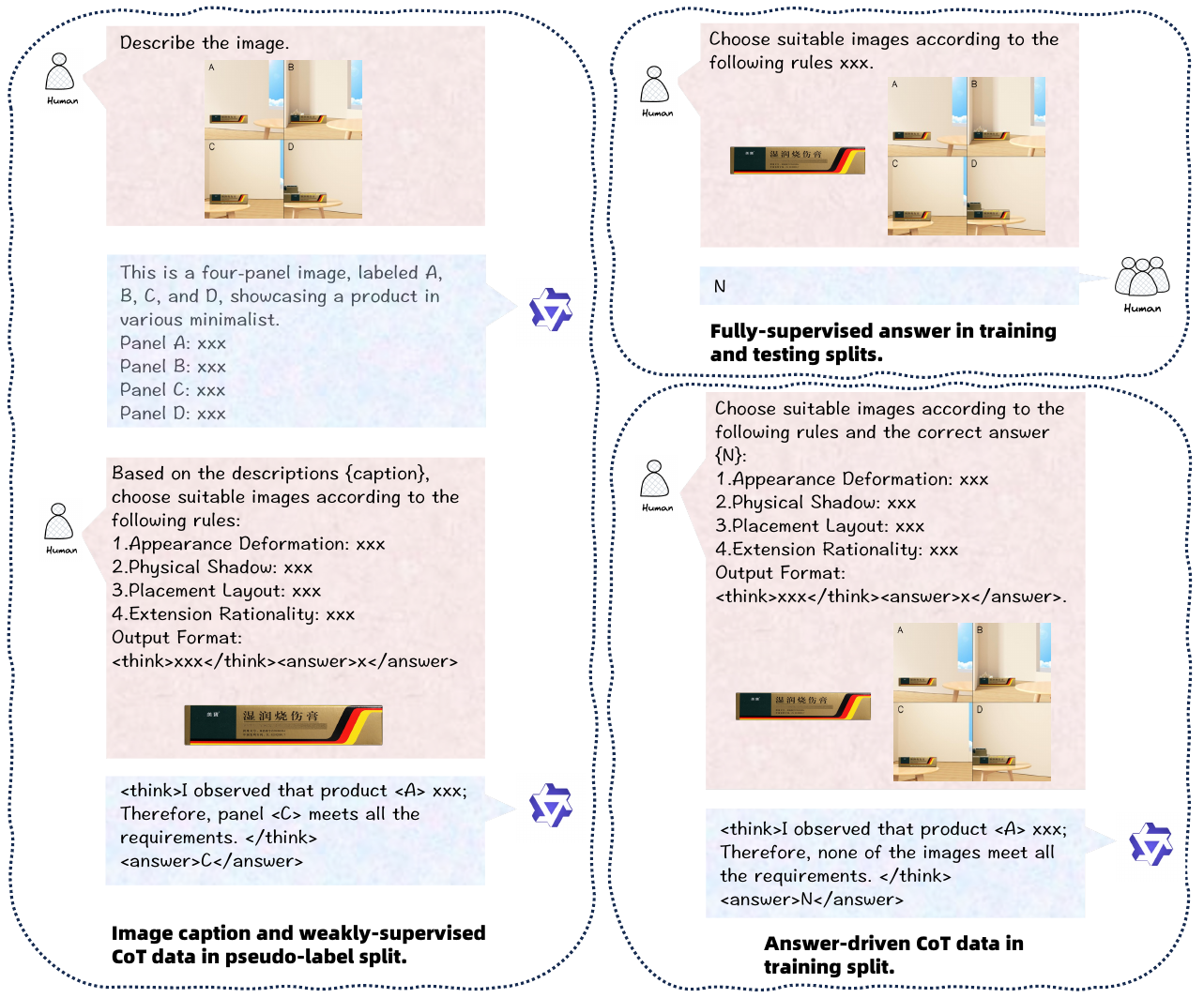}
\caption{Presentation of different annotation paradigms.} \label{data_annotation}
\end{figure}

\subsection{Dataset Division and Annotation}
As shown in Table~\ref{tab:dataset_overview}, our dataset is divided into training, testing, pseudo-label, and exploration splits, comprising a total of 128k unique samples. Each sample includes an original medicine image and four generated images. \\
In the training and testing splits, manual annotation is performed by human reviewers to ensure ground-truth accuracy, as shown in the top-right part of Figure~\ref{data_annotation}. Samples in the training set are labeled with the multi-answer ground-truth, which may contain several correct options, such as "ACD", "BC", or "N". Samples in the testing set contain multi-answer labels for overall evaluation and dedicated evaluation dimension types. This latter annotation allows for a nuanced assessment of the model's capabilities in perceiving different physical plausibility dimensions of the images. The annotation process is exceptionally time-consuming, as it requires a detailed inspection of the four optional images for each sample.\\
Furthermore, annotating CoT data would be even more laborious, as it involves documenting the reasoning process across four distinct physical plausibility dimensions for each of the four options. Given this prohibitive complexity, we leverage the Qwen-VL-Max~\cite{qwen-vl-max} model for automated labeling to enhance the diversity and richness of the descriptive tags in the pseudo-label split. We design a progressive annotation paradigm to enable the model to generate accurate image descriptions and structured reasoning processes. As shown in the left part of Figure~\ref{data_annotation}, for each sample in our pseudo-label split, we first prompt the model to describe what it sees in the image. Inspired by the SIC dataset~\cite{yang2025eye} and MFSC~\cite{fan2026towards}, and considering that  image captioning is a fundamental and common task during the pre-training of MLLMs, we operate under the assumption that the model can generate these basic descriptions with high accuracy. Then, the model is instructed to integrate its own description, structure its reasoning process, and conclude with the final answer. Consequently, within our pseudo-label split, we generate two distinct types of pseudo-labeled data: one is the image description, and the other is the CoT data grounded in this description.\\
For the final exploration split, we generate four synthetic images from each original image but provide no annotations. This portion of the data is intentionally left unlabeled to serve as a testbed for future exploration of unsupervised methods.

% \begin{table}[!ht]
% \centering
% \caption{An overview of the different data splits used, detailing their characteristics such as label accuracy, inclusion of CoT data, and supervision type.}
% \label{tab:dataset_overview}
% \resizebox{\linewidth}{!}{
%     \begin{tabular}{ccccc}
%     \toprule
%     % 使用 \makecell 来允许表头换行
%     \textbf{Data Split} & \makecell{\textbf{1k} \\ \textbf{Training}} & \makecell{\textbf{0.5k} \\ \textbf{Eval}} & \makecell{\textbf{10.5k} \\ \textbf{Pseudo-labels}} & \makecell{\textbf{100k} \\ \textbf{Unlabeled}} \\
%     \midrule
%     Ground-truth Labels & \checkmark & \checkmark & $\times$ & $\times$ \\
%     Includes CoT data   & $\times$ & $\times$ & \checkmark & $\times$ \\
%     Supervision Type    & Fully & Fully & Weakly & Unsupervised \\
%     \bottomrule
%     \end{tabular}
% }
% \end{table}

\section{Method}

\subsection{Cold Start: Basic Spatial Understanding and Instruction Following}
\begin{figure}[!t]
\centering
\includegraphics[width=1.0\textwidth]{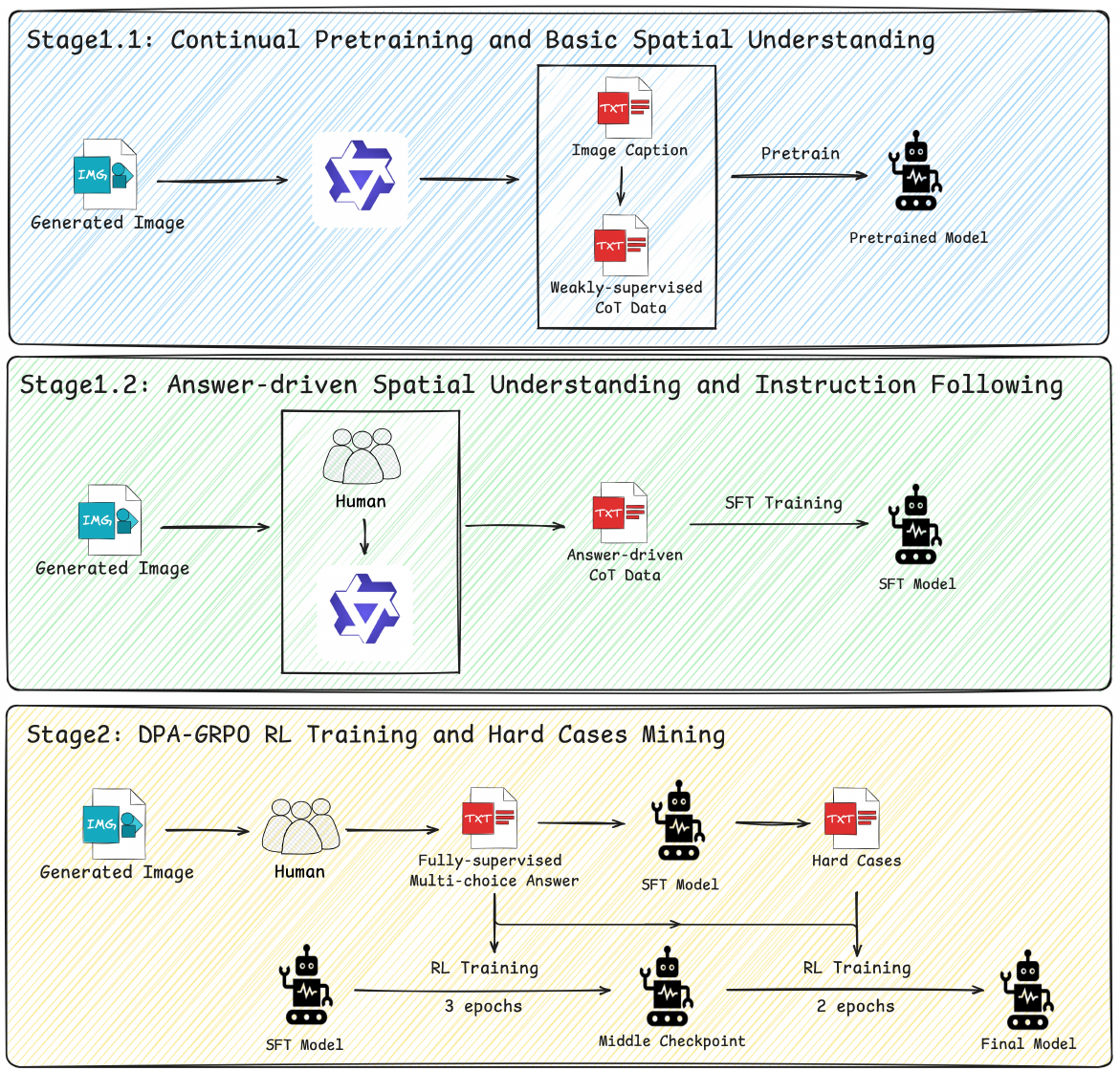}
\caption{Illustration of model training process.} \label{model_training}
\end{figure}

We expect MLLMs to possess basic capabilities in the physical plausibility reasoning task, such as instruction-following and fundamental spatial change recognition. However, the training data for these MLLMs often fails to cover specific scenarios. This limitation results in the base model being unable to output responses in a specific format as instructed, and the generated content often contains hallucinations. As a result, direct application of reinforcement learning methods tends to incur significant consumption of training resources, while the final outcomes still fall short of optimality. Inspired by DeepSeek-R1-Zero and DeepSeek-R1~\cite{guo2025deepseek}, we perform cold-start training using CoT data before the reinforcement learning phase.\\
This process involves acquiring CoT data, which is resource-intensive for the image screening task, as it requires annotators to carefully examine the details of each image. To address this issue, we design two distinct methods for acquiring CoT data. The first method is described in our pseudo-label split, which involves generating an image description followed by the corresponding reasoning steps. Although the accuracy of these weakly-supervised CoT data is relatively modest with an approximate accuracy of 38.25\% (as indicated in Table~\ref{tab:Comparison Results}), our primary focus lies not in the absolute accuracy of the data. Instead, we aim to determine whether this data can enable the model to respond in a fixed format and develop basic spatial change recognition abilities. In our second method, we recognize that the core value of CoT data lies in the reasoning process. Allowing Qwen-VL-Max to directly generate this process would be equivalent to unconstrained self-generation, lacking proper guidance. Therefore, we leverage samples from the training set, conditioning on the human-annotated correct answers, to prompt Qwen-VL-Max to regenerate 1,044 answer-driven CoT data samples. This procedure is illustrated in the bottom-right part of Figure~\ref{data_annotation}. In our training process, these two methods for generating CoT data are applied sequentially. We first perform continual pretraining and basic spatial understanding on the image caption and weakly-supervised CoT data, followed by answer-driven spatial understanding and instruction following on the answer-driven CoT data. This training pipeline is illustrated in Figure~\ref{model_training} Stage1.1 and Stage1.2. The objective function of SFT is defined as:
\begin{equation}
\mathcal{L}_{cold\_start}(\theta)=-\sum_{i=1}^T\log p(y_i|x,y_{<i};\theta),
\end{equation}
where $x$ is the original input, $y=\{y_1,y_2,...,y_T\}$ is the distilled output from Qwen-VL-Max, and $\theta$ represents the parameters of the base model. This stage serves to initialize the model's ability to follow a structured CoT reasoning format.

\subsection{HCM-GRPO: Reinforcement Fine-Tuning and Hard Cases Mining}
GRPO foregoes the critic model that is typically the same size as the policy model, and estimates the baseline from group scores instead~\cite{guo2025deepseek}. Given an input $q$, GRPO first samples $G$ distinct responses $\{o_i\}^G_{i=1}$ from the old policy model $\pi_{old}$. Each response is evaluated by the pre-designed reward functions, i.e., format and accuracy, to calculate the corresponding reward scores $\{r_i\}^G_{i=1}$. Then, GRPO computes the difference between each reward and the mean of the reward score group, which can be regarded as the baseline value, and normalizes the differences using the standard deviation:
\begin{equation}
A_i=\frac{r_i-mean(\{r_1,r_2,...,r_G\})}{std(\{r_1,r_2,...,r_G\})},
\end{equation}
where $A_i$ quantifies the relative quality of the $i$-th response in comparison to other candidates within the same sampled group. Based on the simple advantage $A_i$, GRPO optimizes the policy model $\pi_{\theta}$ by maximizing the following objective:
\begin{equation}
\begin{split}
\mathcal{J}_{GRPO}(\theta) &= \mathbb{E}[q\sim P(Q),\{o_i\}_{i=1}^G\sim\pi_{old}(O|q)] \\
     &\frac{1}{G}\sum_{i=1}^G(min(w_iA_i,clip(w_i,1-\epsilon,1+\epsilon)A_i)-\beta\mathbb{D}_{KL}(\pi_\theta||\pi_{ref})),
\end{split}
\end{equation}

\begin{equation}
\mathbb{D}_{KL}(\pi_{\theta}||\pi_{ref})=\frac{\pi_{ref}(o_i|q)}{\pi_{\theta}(o_i|q)}-log\frac{\pi_{ref}(o_i|q)}{\pi_{\theta}(o_i|q)} -1,
\end{equation}

\begin{equation}
w_i=\frac{\pi_{\theta}(o_i|q)}{\pi_{old}(o_i|q)},
\end{equation}
where $w_i$ is the importance sampling coefficient, $\beta$ is a hyper-parameter that avoids the policy model from diverging too far from the reference model $\pi_{ref}$. $\epsilon$ clips the extreme importance sampling coefficient for stability.\\
In DeepSeek-R1~\cite{guo2025deepseek}, the overall verifiable reward $r$ is formulated as:
\begin{equation}
r=r_{fmt}+r_{acc}, 
\end{equation}
where $r_{fmt}$ and $r_{acc}$ stand for the format reward and the accuracy reward, respectively. The format reward encourages the model's responses to follow a fixed structure, including $<$think$><$/think$><$answer$><$/answer$>$. In this paper, we retain this reward by using prompts, provided in \ref{apd:second}, to guide the model to output responses in a fixed format. Additionally, the role of first-stage CoT learning is to ensure that the model can stably generate responses in this specific format.\\
The accuracy reward evaluates whether the final answer exactly matches the ground truth. Our HCM-GRPO differs from the original GRPO primarily in two aspects: the design of the reward function and the training paradigm. In the task of physical plausibility reasoning, we believe that the design of the accuracy reward for multi-answer questions is crucial. Since a sample may have multiple correct options, we treat the multi-answer label as a complete answer and introduce a Dynamic Proportional Accuracy (DPA) reward. Under this schema, the accuracy reward value is no longer binary, which is tied to the length of the multi-answer labels instead:
% \begin{equation} 
% r_{acc} = 
% \begin{cases} 
% \frac{|\mathcal{R}|}{|\mathcal{A}|}, & \text{if } \mathcal{R} \subseteq \mathcal{A} \\ 
% 0, & \text{otherwise} 
% \end{cases} 
% \end{equation}
\begin{equation} 
r_{acc} = 
\begin{cases} 
\frac{|R|}{|A|}, & \text{if } R \subseteq A(q) \\ 
0, & \text{otherwise} 
\end{cases} 
\end{equation}
where $R$ is the set of chosen options in the response, $A(q)$ is the set of ground-truth answers, and $\vert \cdot \vert$ denotes the number of elements in the set. Crucially, the subset condition $R \subseteq A(q)$ explicitly handles "mixed" outputs: if $R$ contains any incorrect choices, it is no longer a subset of $A(q)$, resulting in a reward of 0. This design grants partial credit only when the response is fully contained within the correct answers, providing a more fine-grained measure of model accuracy. The rationale behind this design is that the reward function should measure the gap between the response and the correct answer as accurately as possible. When humans solve a multi-answer question, a score of 0 should be given if the response is incorrect. However, if the response is part of the correct answer, partial credit should be awarded, with a score between 0 and 1. A full reward, such as 1, is given only when the response is completely correct. We term GRPO with DPA reward as DPA-GRPO. To further enhance the model's sophisticated reasoning abilities beyond the initial supervised fine-tuning, we introduce an advanced reinforcement learning strategy termed Hard Cases Mining in Group Relative Policy Optimization (HCM-GRPO). As illustrated in Figure~\ref{model_training} Stage2, this method is designed to compel the model to learn from its mistakes by focusing on challenging examples. The process begins with a hard case identification step, where we leverage the SFT model from Stage1 to evaluate the entire training set against the human-annotated, fully-supervised multi-answer labels. Samples answered correctly are classified as easy cases, while those answered incorrectly are designated as hard cases. This curated set of hard cases is then integrated into a specialized RL training schedule using the DPA-GRPO algorithm. For the first 3 epochs, the model undergoes general training on the complete dataset to establish a robust baseline. Subsequently, in the final 2 epochs, we implement the hard cases mining strategy by augmenting the training data and re-introducing the identified hard cases alongside the full dataset. This oversampling of challenging examples compels the model to dedicate more learning capacity to its weaknesses, thereby refining its reasoning pathways and leading to a more capable final model.\\
From an RL perspective, the image screening task can be formulated as a contextual set prediction problem, where the policy $\pi_\theta(\cdot|q)$ predicts an answer set $R$ for a query $q$, and the ground-truth is a set of acceptable answers $A(q)$. To provide denser yet well-calibrated learning signals, DPA introduces a normalized partial-credit reward:
\begin{equation}
r_{\mathrm{DPA}}(q,R)
=
\frac{|R \cap A(q)|}{|A(q)|}
\cdot
\mathbf{1}[R \subseteq A(q)],
\end{equation}
where $\mathbf{1}[\cdot]$ is the indicator function that equals $1$ when the condition holds and $0$ otherwise, which preserves the set-level correctness constraint while normalizing rewards across samples with different numbers of valid answers. This formulation can be interpreted as a difficulty-normalized reward shaping strategy that provides more informative feedback for partially correct predictions. Meanwhile, HCM-GRPO adopts a static hard-case partition defined once by the Stage1 SFT model, where the training set is denoted by $\mathcal{D}$, each $q\in\mathcal{D}$ is a training query, and the mispredicted samples form the hard subset $\mathcal{H}\subseteq\mathcal{D}$. The subsequent RL stage can be viewed as optimizing a fixed reweighted objective:
\begin{equation}
J_{\mathrm{HCM-GRPO}}(\theta)
=
\frac{1}{N+\gamma M}
\sum_{q\in\mathcal{D}}
\left(1+\gamma\,\mathbf{1}[q\in\mathcal{H}]\right)
J_{\mathrm{GRPO}}(\theta;q),
\end{equation}
where $\theta$ denotes the policy parameters, $J_{\mathrm{GRPO}}(\theta;q)$ is the per-sample GRPO objective for query $q$, $N=|\mathcal{D}|$ is the size of the full training set, $M=|\mathcal{H}|$ is the number of hard samples, $\gamma\ge0$ controls the extra weight assigned to hard cases, and $\mathbf{1}[\cdot]$ is the indicator function, which equals $1$ when the condition is true and $0$ otherwise. We set $\gamma=1$ and adopt this static design for both stability and efficiency: it keeps the training objective stationary during RL and avoids repeated full-dataset scoring under an evolving policy, which would be computationally expensive in GRPO.\\
% A dynamic variant is a natural extension, where each sample $q$ is assigned a time-dependent weight $w_t(q)$ at iteration $t$, and this weight is updated online according to the current prediction error:
% \begin{equation}
% w_{t+1}(q)=(1-\lambda)w_t(q)+\lambda\big(1+\gamma e_t(q)\big),
% \end{equation}
% where $w_t(q)$ is the sampling weight of query $q$ at iteration $t$, $w_{t+1}(q)$ is the updated weight for the next iteration, $e_t(q)\in\{0,1\}$ indicates whether query $q$ is mispredicted at iteration $t$, $\lambda\in(0,1]$ is a smoothing factor that controls how strongly the new error signal affects the updated weight, and $\gamma\ge0$ again determines the additional emphasis placed on hard samples. Such a strategy is related to prioritized replay or self-paced learning and may better track emerging failure modes, but it also introduces additional computation and potential instability due to a moving hard set. Given the limited RL budget and the strong empirical gains of the stationary schedule, we adopt static HCM as a simple and stable approximation to prioritized training.
\begin{table}[t]
    \centering
    \renewcommand{\arraystretch}{0.85} 
    % \small
    \caption{Training parameters and computational resources. We compare several distinct training paradigms, encompassing experimental hyperparameters and computational resources. Notably, the training hours listed are standardized using the InternVL3-2B model trained on the training split for a fair comparison.}
    \label{tab:Training Parameters}
    \resizebox{\linewidth}{!}{
    \begin{tabular}{c c c c}
        \toprule
        \textbf{Parameter} & \textbf{SFT} & \textbf{GRPO (DPA-GRPO)} & \textbf{HCM-GRPO}\\
        \midrule
        Freeze LLM & False & False & False\\
        Freeze ViT & True & False & False\\
        Freeze Aligner & True & False & False\\
        Epoch & 5 & 5 & 3+2\\
        Per-device Batch Size & 8 & 8 & 8\\
        $G$ for Group Generations & NULL & 4 & 4\\
        $\beta$ for KL divergence & NULL & 0.04 & 0.04 \\
        $Top_k$ & NULL & 3 & 3\\
        $Top_p$ & NULL & 0.8 & 0.8\\
        $Temperature$ & NULL & 1 & 1\\
        Learning Rate & 1e-4 & 1e-5 & 1e-5 \\
        Optimizer & AdamW & AdamW & AdamW\\
        GPU (PPU-ZW810E) Number & 8 & 8 & 8\\
        Training Hour & $\approx 0.5$ & $\approx 10.6$ & $\approx 12.7$\\
        \bottomrule
    \end{tabular}}
\end{table}

\section{Experimental Result}

\subsection{Experimental Setting}
We conduct experiments on the collected image screening dataset for the task of physical plausibility reasoning. Furthermore, we validate its performance on several public datasets to assess its capabilities in real-world spatial reasoning and multi-image understanding. Our HCM-GRPO demonstrates competitive performance by utilizing hard cases mining in GRPO compared to others.\\
\textbf{Baseline Models.} For the baseline models used for cross-sectional comparison, we select several large-sized open-source models, such as InternVL3.5-30B-A3B~\cite{wang2025internvl3_5}, InternVL3-38B~\cite{chen2024internvl}, Qwen3.5-397B-A17B~\cite{qwen35blog}, Qwen2.5-VL-72B-Instruct~\cite{qwen2.5-VL}, and GLM-4.5V~\cite{hong2025glm}, as well as some closed-source models like GPT5.2~\cite{openai-gpt}, GPT4o~\cite{openai-gpt}, Gemini3-Pro~\cite{google-gemini3}, ClaudeSonnet4.5~\cite{anthropic-claude-sonnet-4-5}, Qwen3.5-Plus~\cite{qwen35blog}, and Qwen-VL-Max~\cite{qwen-vl-max}. For the baseline compact models used for longitudinal comparison, we select InternVL3-1B, InternVL3-2B~\cite{chen2024internvl}, and Qwen2.5-VL-3B~\cite{qwen2.5-VL} to validate the effectiveness of our proposed HCM-GRPO.\\
\textbf{Hyperparameter Settings and Computational Resources.} All experiments are implemented via trl based on the training framework SWIFT~\cite{zhao2025swift}, a flexible and efficient tool for supervised fine-tuning and reinforcement learning. As shown in Table~\ref{tab:Training Parameters}, we compare several distinct training paradigms. In SFT, the Vision Transformer (ViT) and Aligner of MLLMs are frozen, with no generative sampling strategies, focusing on structured supervised learning with a batch size of 8. In contrast, GRPO unfreezes all components and introduces advanced generative sampling techniques, including $Top_k=3$, $Top_p=0.8$, and $Temperature=1$, enabling stochasticity during training. Additionally, GRPO incorporates a KL divergence penalty with a $\beta$ coefficient of 0.04 to balance exploration and stability, while SFT omits this term entirely. Despite sharing common settings, such as 5 training epochs, the two paradigms differ significantly in their designs. GRPO highlights its adaptability for complex tasks, whereas SFT emphasizes simplicity and stability. In terms of training time, while HCM-GRPO requires about 2 hours more than GRPO, this overhead is considered negligible relative to the significant cost increase from SFT (0.5 hours) to GRPO (10.6 hours), making it an acceptable trade-off.\\
\textbf{Evaluation Metric.} Due to the complexity of physical plausibility reasoning, we primarily employ a task success metric (Overall Score) tailored to our deployment scenario. Specifically, for multi-answer questions, a model response is considered correct if it constitutes a subset of the ground-truth options. This design aligns with our real-world e-commerce task: product image screening. In this context, only one image is ultimately displayed on the webpage; therefore, as long as the model recommends a subset of ground-truth options, the task is deemed successful from the user's perspective. Consequently, the task success metric is the most appropriate primary indicator for this application, as it considers a prediction successful when all selected recommendations are correct, even if the model returns only a subset of the full set of valid answers. Unless otherwise specified, all subsequent experiments utilize this metric.\\
% \textbf{Random Guessing.} The random guessing baseline score of 34.27 is a calculated expectation rather than an empirical result from a random simulation. It is derived from the specific structure of our test set, where each sample can have one or more correct options. Through statistical analysis of the test set, we determine that the average length of correct options per sample is 1.7136. Since a guess involves picking one from five choices (A, B, C, D, N), the expected probability of success is 1.7136 divided by 5, which equals 0.3427.\\
\begin{table}[!t]
    \centering
    \renewcommand{\arraystretch}{0.85} 
    % \small
    \caption{Comparison results. We evaluate the image screening performance of both closed-source and open-source MLLMs. We use bold to highlight the top results, and underline to indicate the second-best results. Notably, the overall score includes four dimensions of physical plausibility ability and the normal type.}
    \label{tab:Comparison Results}
    % 默认是 1.0，设置为 0.85 可以减少行与行之间的留白
    \renewcommand{\arraystretch}{0.9} 
    % 如果还需要减少列间距，可以取消下面这行的注释
    %\setlength{\tabcolsep}{1pt} % 减少列间距
    %\footnotesize
    \resizebox{\linewidth}{!}{
        \begin{tabular}{c c | c | c c c c}
            \toprule
            \textbf{Models} & 
            \makecell{\textbf{Prompt} \\ \textbf{Type}} & 
            \makecell{\textbf{Overall} \\ \textbf{Score}} &
            \makecell{\textbf{Appearance} \\ \textbf{Deformation}} &
            \makecell{\textbf{Physical} \\ \textbf{Shadow}} &
            \makecell{\textbf{Placement} \\ \textbf{Layout}} &
            \makecell{\textbf{Extension} \\ \textbf{Rationality}} \\
            \midrule
            % Random & - & 34.27 & 27.61 & 31.46 & 29.49 & 27.25\\
            % \midrule
            \multicolumn{7}{c}{\textbf{API-based models}}\\
            \midrule
            % \multirow{2}{*}{GPT-5-0807-global} & Direct & x & x & x & x & x\\
            %                               & CoT & x & x & x & x & x\\
            \multirow{2}{*}{GPT5.2} & Direct & 50.00 & 46.24 & 51.22 & 53.39 & 49.50\\
                                   & CoT    & 42.95 & 32.09 & 45.12 & 33.90 & 44.55\\
            \multirow{2}{*}{GPT4o} & Direct & 18.59 & 18.66 & 13.41 & 22.88 & 20.59\\
                                   & CoT    & 34.62 & 23.51 & 29.27 & 24.58 & 29.41\\
            \multirow{2}{*}{Gemini3-Pro} & Direct & 44.02 & 33.58 & 39.02 & 37.29 & 33.33\\
                                          & CoT & 37.39 & 27.61 & 34.15 & 26.27 & 25.49\\
            \multirow{2}{*}{ClaudeSonnet4.5} & Direct & 40.17 & 29.59 & 35.80 & 28.45 & 43.14\\
                                          & CoT & 37.82 & 26.59 & 34.15 & 23.73 & 39.22\\
            \multirow{2}{*}{Qwen3.5-Plus} & Direct & 43.59 & 34.33 & 40.24 & 33.90 & 37.25\\
                                         & CoT & 39.74 & 30.60 & 39.02 & 27.97 & 34.31\\
            \multirow{2}{*}{Qwen-VL-Max} & Direct & 32.05 & 21.64 & 26.83 & 25.42 & 15.67\\
                                         & CoT & 38.25 & 29.10 & 37.80 & 25.42 & 28.43\\
            % \multirow{2}{*}{Qwen-VL-Plus} & Direct & 34.62 & 22.76 & 35.37 & 25.42 & 27.45\\
            %                               & CoT & 32.26 & 21.27 & 29.27 & 25.42 & 28.43\\
            \midrule
            \multicolumn{7}{c}{\textbf{Open-source MLLMs}}\\
            \midrule
            \multirow{2}{*}{GLM-4.5V} & Direct & 34.83 & 23.88 & 29.27 & 27.97 & 26.47\\
                                          & CoT & 32.69 & 20.90 & 30.49 & 22.03 & 23.53\\
            % \multirow{2}{*}{Qwen3-VL-235B-A22B-Instruct} & Direct & 37.82 & 26.49 & 35.37 & 30.51 & 25.49\\
            %                               & CoT & x & x & x & x & x\\
            \multirow{2}{*}{Qwen3.5-397B-A17B} & Direct & 43.80 & 35.82 & 37.80 & 33.05 & 39.22\\
                                          & CoT & 39.96 & 31.34 & 35.37 & 31.36 & 32.35\\
            \multirow{2}{*}{Qwen3-VL-30B-A3B-Instruct} & Direct & 33.97 & 23.51 & 25.61 & 25.42 & 26.47\\
                                          & CoT & 36.54 & 27.99 & 36.59 & 22.88 & 30.39\\
            \multirow{2}{*}{Qwen3-VL-8B-Instruct} & Direct & 34.62 & 23.88 & 32.93 & 24.58 & 24.51\\
                                          & CoT & 34.40 & 21.64 & 29.27 & 25.42 & 24.51\\
            % \multirow{2}{*}{Qwen3-VL-4B-Instruct} & Direct & x & x & x & x & x\\
            %                               & CoT & x & x & x & x & x\\
            \multirow{2}{*}{Qwen2.5-VL-72B-Instruct} & Direct & 39.74 & 29.85 & 40.24 & 32.20 & 30.39\\
                                          & CoT & 35.47 & 25.00 & 35.37 & 25.42 & 25.49\\
            % \multirow{2}{*}{InternVL3.5-241B-A28B} & Direct & x & x & x & x & x\\
            %                               & CoT & x & x & x & x & x\\
            \multirow{2}{*}{InternVL3.5-30B-A3B} & Direct & 32.48 & 20.90 & 26.83 & 22.03 & 26.47\\
                                          & CoT & 27.56 & 18.28 & 21.95 & 22.03 & 25.49\\
            \multirow{2}{*}{InternVL3.5-38B} & Direct & 33.33 & 20.90 & 26.83 & 25.42 & 25.49\\
                                          & CoT & 20.73 & 13.06 & 21.95 & 13.56 & 15.69\\
            % \multirow{2}{*}{InternVL3-78B} & Direct & 34.83 & 23.88 & 31.71 & 24.58 & 27.45\\
            %                               & CoT & 36.75 & 26.49 & 31.71 & 29.66 & 28.43\\
            \multirow{2}{*}{InternVL3-38B} & Direct & 36.75 & 26.49 & 32.93 & 30.51 & 27.45\\
                                          & CoT & 32.48 & 21.27 & 24.39 & 24.58 & 26.47\\
            \midrule
            HCM-GRPO-1B (Ours) & CoT & 57.48 & 58.21 & 60.98 & \underline{62.71} & 57.84\\
            HCM-GRPO-2B (Ours) & CoT & \underline{64.74} & \underline{68.66} & \underline{65.85} & 55.08 & \underline{74.51}\\
            HCM-GRPO-3B (Ours) & CoT & \textbf{71.15} & \textbf{74.25} & \textbf{69.51} & \textbf{63.56} & \textbf{81.37}\\
            \bottomrule
        \end{tabular}
    }
\end{table}

\subsection{Comparison with Prior Works}
We compare our proposed HCM-GRPO method on small models with several open-source large-sized and closed-source MLLMs. During evaluation, we assess the performance of both direct and CoT format prompts, reporting the overall score as well as the scores for the four evaluation dimensions. The results are shown in Table~\ref{tab:Comparison Results}. In general, all models exhibit suboptimal performance on our dataset. This indicates that these models lack physical plausibility reasoning capabilities. In this comparative analysis, closed-source models exhibit superior performance relative to open-source alternatives. Notably, GPT5.2 stands out as the most proficient model.\\
For our own HCM-GRPO method, we apply HCM-GRPO to different compact models, such as InternVL3-1B, InternVL3-2B, and Qwen2.5-VL-3B, obtaining HCM-GRPO-1B, HCM-GRPO-2B, and HCM-GRPO-3B. As shown in Table~\ref{tab:Comparison Results}, our models achieve substantial improvements across most dimensions, particularly in appearance deformation and extension rationality, which demonstrates that the HCM-GRPO method can generalize across architectures, not just model sizes within the same family. However, it still struggles with recognizing more advanced physical rules, such as placement layout. The experimental results align with human-like recognition abilities: changes in appearance and quantity are indeed easier to comprehend compared to physical layout and shadow. Some reasoning cases are illustrated in Figure~\ref{Demonstration2}.

\begin{table}[t]
    \centering
    \renewcommand{\arraystretch}{0.85} 
    % \small
    \caption{Comparison between direct answering and reasoning.}
    \label{tab:abation_Comparison}
    \begin{tabular}{c c c}
        \toprule
        \textbf{Models} & \textbf{Prompt Type} & \textbf{Overall Score}\\
        \midrule
        \multirow{2}{*}{InternVL3-1B} & Direct & 30.13\\
                                      & CoT    & 25.85\\
        +SFT & Direct & \textbf{36.75}\\
        +GRPO & CoT & 28.42\\
        \midrule
        \multirow{2}{*}{InternVL3-2B} & Direct & 10.68\\
                                      & CoT    & 23.72\\
        +SFT & Direct & \textbf{45.94}\\
        +GRPO & CoT & 28.42\\
        \bottomrule
    \end{tabular}
\end{table}

\begin{table}[t]
    \centering
    \renewcommand{\arraystretch}{0.85} 
    % \small
    \caption{Effectiveness of different CoT data with InternVL3-1B and InternVL3-2B.}
    \label{tab:abation_Effectiveness}
    \resizebox{\linewidth}{!}{
    \begin{tabular}{c c c c c}
        \toprule
        \textbf{Exp} & 
        \makecell{\textbf{Image Caption} \\ \textbf{Data}} &
        \makecell{\textbf{Weakly-supervised} \\ \textbf{CoT Data}} &
        \makecell{\textbf{Answer-driven} \\ \textbf{CoT Data}} &
        \textbf{Overall Score}\\
        % \midrule
        % \multirow{2}{*}{InternVL3-1B} & Direct & 30.13\\
        %                               & CoT    & 25.85\\
        % +SFT & Direct & \textbf{36.75}\\
        % +GRPO & CoT & 28.42\\
        \midrule
        InternVL3-1B &  &  &  & 25.85\\
        Exp1 & $\times$ & $\checkmark$ & $\times$ & 34.19 (+8.34)\\
        Exp2 & $\checkmark$ & $\checkmark$ & $\times$ & 34.62 (+8.77)\\
        Exp3 & $\times$ & $\times$ & $\checkmark$ & 39.96 (+14.11)\\
        Exp4 & $\checkmark$ & $\checkmark$ & $\checkmark$ & \textbf{45.94 (+20.09)}\\
        \midrule
        InternVL3-2B &  &  &  & 23.72\\
        Exp1 & $\times$ & $\checkmark$ & $\times$ & 34.83 (+11.11)\\
        Exp2 & $\checkmark$ & $\checkmark$ & $\times$ & 35.04 (+11.32)\\
        Exp3 & $\times$ & $\times$ & $\checkmark$ & 51.07 (+27.35)\\
        Exp4 & $\checkmark$ & $\checkmark$ & $\checkmark$ & \textbf{53.63 (+29.91)}\\
        \bottomrule
    \end{tabular}}
\end{table}

\begin{table}[t]
    \centering
    \renewcommand{\arraystretch}{0.85} 
    % \small
    \caption{Comparison among different GRPO paradigms and Overall Score* represents the exact match metric.
    % InternVL3-1B-CoT* means performing SFT with only answer-driven CoT data.
    }
    \label{tab:abation_Comparison_GRPO}
    \begin{tabular}{c c c}
        \toprule
        \textbf{Models} & \textbf{Overall Score} & \textbf{Overall Score*}\\
        \midrule
        InternVL3-1B-CoT & 45.94 & 19.65\\
        +GRPO & 53.85 (+7.91) & 38.46 (+18.81)\\
        +DPA-GRPO & 55.56 (+9.62) & 40.38 (+20.73)\\
        +HCM-GRPO & \textbf{57.48 (+11.54)} & \textbf{41.88 (+22.23)}\\
        \midrule
        InternVL3-2B-CoT & 53.63 & 23.72\\
        % +Ori Acc \& DPA & 52.56 (+13.67)\\
        +GRPO & 58.55 (+4.92) & 38.89 (+15.17)\\
        +DPA-GRPO & 59.83 (+6.20) & 41.45 (+17.73)\\
        +HCM-GRPO & \textbf{64.74 (+11.11)} & \textbf{44.66 (+20.94)}\\
        \bottomrule
    \end{tabular}
\end{table}

\subsection{Ablation Studies}
We conduct a series of experiments to evaluate the effectiveness of different modules in our approach based on InternVL3-1B, InternVL3-2B, and Qwen2.5-VL-3B.\\
\textbf{Comparison Between Direct Answering and Reasoning.} By using the training set, we compare the original models, models fine-tuned with direct answers, and models trained with GRPO. As shown in Table~\ref{tab:abation_Comparison}, direct answer fine-tuning achieves the highest scores (36.75 and 45.94). In contrast, applying GRPO directly to the base models is ineffective. This likely occurs because the original models lack sufficient instruction-following ability and basic spatial understanding, causing the accuracy-based reward to fail.\\
\textbf{Effectiveness of Different Sources of CoT Data.} Table~\ref{tab:abation_Comparison} shows that direct answer fine-tuning yields a good improvement, while RL methods such as GRPO are ineffective on the base model. This motivates building a base model with basic reasoning ability first. To this end, we use collected CoT data, including image captions, weakly-supervised CoT generated by Qwen-VL-Max on pseudo-label data, and human-annotated answer-driven CoT in the training set.
The ablation study in Table~\ref{tab:abation_Effectiveness} analyzes different CoT acquisition strategies. On InternVL3-2B, weakly-supervised CoT alone (Exp1) improves performance from 23.72 to 34.83, indicating that auto-generated reasoning steps provide a useful foundation. Adding image captions (Exp2) brings only marginal gains. In contrast, answer-driven CoT alone (Exp3) significantly boosts performance to 51.07, highlighting the value of high-quality human-guided reasoning. Combining all data sources (Exp4) achieves the best results, slightly surpassing Exp3, suggesting that captions and weak supervision can further complement high-quality CoT. Similar trends are observed on InternVL3-1B.\\
\textbf{Comparison among Different GRPO Paradigms.} From Table~\ref{tab:abation_Comparison}, directly applying GRPO to the original models is ineffective. Therefore, we start from the CoT-initialized models in Table~\ref{tab:abation_Effectiveness}, namely InternVL3-1B-CoT and InternVL3-2B-CoT, obtained via SFT with different CoT sources. We then compare different reward designs, including the original binary accuracy reward and the proposed DPA reward, while keeping the same CoT prompt. Results in Table~\ref{tab:abation_Comparison_GRPO} show that reward design critically affects RL effectiveness. The original accuracy reward performs worse than the proposed DPA reward, indicating that rewards must better reflect the intended actions. For the multi-answer task in this work, DPA aligns more closely with the actual scoring rules. Furthermore, integrating hard case mining into DPA-GRPO forms HCM-GRPO, which achieves the best score of 64.74 on InternVL3-2B-CoT, demonstrating that hard case mining further improves the model’s performance ceiling.
For InternVL3-1B-CoT, we further compare GRPO, DPA-GRPO, and HCM-GRPO, obtaining conclusions consistent with those on InternVL3-2B-CoT, which confirms the effectiveness of HCM-GRPO. To address potential metric leniency, we also evaluate using a stricter exact match metric (Overall Score*), which requires identifying all correct options. As shown in Table~\ref{tab:abation_Comparison_GRPO}, although absolute scores drop under this metric, DPA-GRPO and HCM-GRPO consistently outperform the baselines, demonstrating that our method improves reasoning completeness rather than merely identifying partial answers.\\
\textbf{Disentangling the Effects of SFT Data and HCM-GRPO.} To isolate the benefits of the RL process, we conduct a controlled comparison on the InternVL3-2B-CoT and Qwen2.5-VL-3B-CoT, evaluating HCM-GRPO against SFT utilizing the identical answer-driven CoT data (including an extra 434 hard cases). As shown in Table~\ref{tab:abation_Disentangling}, when we further fine-tune InternVL3-2B-CoT and Qwen2.5-VL-3B-CoT with exactly the same answer-driven CoT supervision data (augmented with 434 hard cases), the gains from pure SFT remain marginal: InternVL3-2B-CoT only improves from 53.63 to 55.13 (+1.50), and Qwen2.5-VL-3B-CoT increases from 58.76 to 60.47 (+1.71). 
\begin{table}[t]
    \centering
    \renewcommand{\arraystretch}{0.85} 
    % \small
    \caption{Disentangling the effects of SFT data (including hard cases) and HCM-GRPO.
    % InternVL3-1B-CoT* means performing SFT with only answer-driven CoT data.
    }
    \label{tab:abation_Disentangling}
    \begin{tabular}{c c}
        \toprule
        \textbf{Models} & \textbf{Overall Score}\\
        \midrule
        InternVL3-2B-CoT & 53.63\\
        +SFT & 55.13 (+1.50)\\
        +HCM-GRPO & \textbf{64.74 (+11.11)}\\
        \midrule
        Qwen2.5-VL-3B-CoT & 58.76\\
        +SFT & 60.47 (+1.71)\\
        +HCM-GRPO & \textbf{71.15 (+12.39)}\\
        \bottomrule
    \end{tabular}
\end{table}
In contrast, replacing SFT with HCM-GRPO on the same data yields a dramatic leap: InternVL3-2B-CoT reaches 64.74 (+11.11), and Qwen2.5-VL-3B-CoT climbs to 71.15 (+12.39). More importantly, the hard cases become substantially more effective under RL, indicating that HCM-GRPO can further activate the learning potential hidden in difficult samples rather than being bottlenecked by imitation-style training.\\
\textbf{Different Types of Mining Cases in HCM-GRPO.} To investigate the impact of different data augmentation strategies during our HCM-GRPO training process, we conduct a comprehensive ablation study. The core of this study focuses on the composition of the dataset used in the last 2 epochs of the training stage. All experimental setups share a common first stage: the model is initially trained for 3 epochs on the fully-supervised training set of 1,044 multi-answer samples under the DPA-GRPO framework. Subsequently, we continue training for an additional 2 epochs, but vary the data composition to analyze its effect on final performance. To ensure robustness, all experiments are repeated 5 times, and results are reported as mean ± standard deviation. The results in Table~\ref{tab:abation_Last} demonstrate the effectiveness of our data strategy. For InternVL3-2B-CoT, training on a mixture of the original 1,044 samples and curated hard cases (HCM-GRPO) achieves the best score of 64.87 ± 0.90, which is statistically significant compared with the DPA-GRPO baseline (59.87 ± 0.41, paired t-test). This also surpasses random augmentation, indicating that the improvement comes from targeted hard cases rather than increased data volume.
We explicitly monitor the final HCM-GRPO-2B's performances separately on easy and hard subsets. We observe that the final model maintains stable accuracy on easy cases (typically 95\%+), and the gains brought by the RL stage are primarily concentrated on the hard cases rather than coming at the expense of easy ones. Training only on hard or easy subsets yields worse results than the baseline. Overall, the optimal strategy is to supplement the fully supervised dataset with curated hard cases, enabling the model to address its weaknesses while preserving the overall data distribution.
% This suggests two key points: (1) focusing only on difficult examples may cause the model to overfit to these niche cases and forget the general data distribution; (2) reinforcing only what the model already knows is an inefficient strategy that fails to address its deficiencies.

\begin{table}[!t]
\centering
\renewcommand{\arraystretch}{0.85} 
% \small
\caption{Ablation study on the composition of training data in the last 2 epochs of stage2. All models are initially trained on 1,044 supervised samples for first 3 epochs and last 2 epochs with the specified data. Results are reported as mean ± std dev over 5 runs. * indicates statistical significance (p < 0.05) compared to the DPA-GRPO baseline.}
\label{tab:abation_Last}
\begin{tabular}{c c c}
\toprule
\textbf{Last 2 Epochs Data Composition} & \textbf{Overall Score} & \textbf{P-value}\\
% \midrule
% \multicolumn{2}{c}{\textbf{InternVL3-1B-CoT}}\\
% \midrule
% All 1044 Samples + xxx Hard Cases (HCM-GRPO)       & \textbf{xx} \\
% All 1044 Samples + xxx Random Samples   & xx \\
% All 1044 Samples (DPA-GRPO)          & xx\\
% xxx Easy Cases Only                  & xx \\
% xxx Hard Cases Only                  & xx \\
\midrule
\multicolumn{3}{c}{\textbf{InternVL3-2B-CoT}}\\
\midrule
434 Hard Cases Only                  & 58.72 (±0.72)* & 0.0142\\
610 Easy Cases Only                  & 59.40 (±0.34) & 0.0825\\
All 1,044 Samples (DPA-GRPO)          & 59.87 (±0.41) & -\\
All 1,044 Samples + 434 Random Samples   & 60.81 (±0.54)* & 0.0147\\
All 1,044 Samples + 434 Hard Cases (HCM-GRPO)       & \textbf{64.87 (±0.90)*} & 0.0001\\
\bottomrule
\end{tabular}
\end{table}

\subsection{Generalization of DPA-GRPO on Public Multi-answer Datasets}
We further evaluate DPA-GRPO on public datasets targeting more general multi-answer scenarios, including Mmlu-Multi-Answer~\cite{mmlu-multi} and JEC-QA~\cite{zhong2024agieval}. The Mmlu-Multi-Answer contains 3,362 general multi-answer instances, and the JEC-QA includes 1,999 multi-answer samples. In each dataset, we split it into training and testing sets at a 1:1 ratio and evaluate InternVL3-2B with different fine-tuning methods. As shown in Table~\ref{tab:generalization_dpa}, DPA-GRPO outperforms GRPO by 15.82 points, achieving a score of 61.45 on the Mmlu-Multi-Answer dataset. For this dataset, we conclude that in more general multi-answer scenarios, the DPA reward is more aligned with the scoring criteria, while the sparse reward signal of GRPO is not suitable. Furthermore, the experimental results from the JEC-QA dataset also support this conclusion, confirming the generality of DPA-GRPO in multi-answer scenarios. The smaller margin of improvement for DPA-GRPO over GRPO on the JEC-QA dataset can be attributed to its data composition: JEC-QA only partially contains multi-answer questions, while Mmlu-Multi-Answer is exclusively composed of them.

\begin{table}[!t]
\centering
\renewcommand{\arraystretch}{0.85}
\setlength{\tabcolsep}{4pt}
\caption{Generalization study on the multi-answer public datasets.}
\label{tab:generalization_dpa}
\begin{tabular}{c c c}
\toprule
\textbf{model} & \textbf{Mmlu-Multi-Answer} & \textbf{JEC-QA}\\
\midrule
InternVL3-2B    & 24.63 & 50.70\\
+SFT            & 34.21 (+9.58) & 59.90 (+9.20)\\
+SFT, +GRPO     & 45.63 (+21.00) & 63.40 (+12.70)\\
+SFT, +DPA-GRPO & \textbf{61.45 (+36.82)} & \textbf{64.20 (+13.50)}\\
\bottomrule
\end{tabular}
\end{table}

\subsection{Generalization of HCM-GRPO on Public Physical and Multi-image Benchmarks}
Question answering has been widely adopted as a paradigm for evaluating reasoning capabilities in large models, as demonstrated by works such as PLRTQA~\cite{alawwad2025enhancing}. Considering the composition of our physical plausibility dataset, where each instance comprises multiple generated images for the purpose of identifying physical space errors, further evaluation on other public datasets is conducted. These public datasets were chosen primarily to evaluate HCM-GRPO in real-world physical reasoning and multi-image collaborative understanding. We select RealWorldQA~\cite{realwordqa}, MuirBench~\cite{wang2024muirbench}, and BLINK~\cite{fu2024blink} for further evaluation.\\
\begin{table}[!t]
\centering
\renewcommand{\arraystretch}{0.85} 
% \small
\caption{Generalization study on the public dataset RealWorldQA.}
\label{tab:generalization1}
\begin{tabular}{c c}
\toprule
\textbf{model} & \textbf{Overall Accuracy}\\
\midrule
InternVL3-2B    & 61.52\\
+SFT            & 62.57 (+1.05)\\
+SFT, +GRPO           & 62.83 (+1.31)\\
+SFT, +HCM-GRPO       & \textbf{64.14 (+2.62)}\\
\bottomrule
\end{tabular}
\end{table}
The RealWorldQA~\cite{realwordqa} dataset comprises 765 instances from the real physical world, with each instance consisting of a question, an image, and its corresponding answer. We partition the RealWorldQA dataset into training and testing sets at a 1:1 ratio. On the testing set, we evaluate the performance differences among various fine-tuning methods, including SFT, GRPO, and our proposed HCM-GRPO, applied to the InternVL3-2B model. As shown in Table~\ref{tab:generalization1}, the InternVL3-2B model achieves a score of 61.52. Our proposed HCM-GRPO combined with SFT outperforms GRPO, demonstrating its effectiveness in enhancing the model's understanding of real-world physical spaces. For this application, HCM-GRPO is configured as the GRPO framework augmented with hard cases mining, excluding the DPA reward component due to the single option nature of the RealWorldQA dataset. This specific implementation, despite lacking complex fine-tuning stages such as answer-driven CoT data, still demonstrates a notable performance gain of 2.62 points relative to the baseline model.\\
% The MIRB~\cite{zhao2024benchmarking} dataset assesses multi-image understanding of MLLMs across four primary domains: multi-image reasoning, visual world knowledge, perception, and multi-hop reasoning. It contains a total of 969 instances (including plot code understanding), each composed of a question, an answer, and several associated images. We partition the data from all subtasks within the MIRB dataset into a 1:1 split for training and testing. Following the same experimental protocol as with RealWorldQA, we then evaluate and compare the performance of the baseline InternVL3-2B model against versions fine-tuned with SFT, GRPO, and our proposed HCM-GRPO on the newly created testing split.\\
% \begin{table}[!h]
% \centering
% \caption{Generalization study on the public dataset MIRB.}
% \label{tab:generalization1}
% \begin{tabular}{c c}
% \toprule
% \textbf{model} & \textbf{Overall Accuracy}\\
% \midrule
% InternVL3-2B    & 26.71\\
% +SFT            & 33.12 (+6.41)\\
% +GRPO           & xx\\
% +HCM-GRPO       & \textbf{xx}\\
% \bottomrule
% \end{tabular}
% \end{table}
\begin{table}[!t]
\centering
\renewcommand{\arraystretch}{0.85} 
% \small
\caption{Generalization study on the public dataset MuirBench.}
\label{tab:generalization2}
\begin{tabular}{c c}
\toprule
\textbf{model} & \textbf{Overall Accuracy}\\
\midrule
InternVL3-1B    & 29.48\\
+SFT            & 62.35 (+32.87)\\
+SFT, +GRPO           & 63.04 (+33.56)\\
+SFT, +HCM-GRPO       & \textbf{63.43 (+33.95)}\\
\bottomrule
\end{tabular}
\end{table}
MuirBench~\cite{wang2024muirbench} is a benchmark containing 11,264 images and 2,600 multi-choice questions, providing robust evaluation on 12 multi-image understanding tasks. We partition the data from all subtasks within the MuirBench dataset into a 1:1 split for training and testing, following the same experimental protocol as with RealWorldQA. Since some samples in the dataset contain a large number of images, which may lead to excessive GPU memory consumption during training, we choose InternVL3-1B as our baseline for validation. We then evaluate and compare the performance of the baseline against versions fine-tuned with SFT, GRPO, and our proposed HCM-GRPO on the newly created testing split. We observe that MuirBench exhibits severe visual and semantic homogeneity, which is consistent with its construction based on minimally different image pairs and high-overlap relations, such as cropped or zoomed variants. As a result, SFT already achieves a large performance gain on this benchmark, leaving only a small proportion of hard cases (15.3\% of the training set), and thus limiting the additional improvement achievable by HCM-GRPO. Even under these conditions, our HCM-GRPO method demonstrates its effectiveness by delivering the highest performance of 63.43.\\
The BLINK dataset consists of 14 classic computer vision tasks, ranging from low-level pattern matching to mid-level reasoning and extending to high-level visual understanding. This benchmark contains 3.8K samples, in which questions may contain multiple images. Since the test set of the BLINK dataset is unlabeled, we conduct our experiments on the validation set with InternVL3-1B and InternVL3-2B. The results in Table~\ref{tab:generalization3} demonstrate the effectiveness of our proposed HCM-GRPO on the BLINK dataset. Both GRPO and HCM-GRPO provide better performance for the InternVL3-1B and InternVL3-2B models. The combination of SFT and HCM-GRPO proves most effective, boosting the accuracy of InternVL3-2B to 49.10 and showcasing a substantial improvement of 8.92 over the base model.\\

\begin{table}[!t]
\centering
\renewcommand{\arraystretch}{0.85} 
% \small
\caption{Generalization study on the public dataset BLINK.}
\label{tab:generalization3}
\begin{tabular}{c c}
\toprule
\textbf{model} & \textbf{Overall Accuracy}\\
\midrule
InternVL3-1B    & 33.30\\
+SFT            & 35.78 (+2.48)\\
+SFT, +GRPO           & 36.79 (+3.49)\\
+SFT, +HCM-GRPO       & \textbf{38.15 (+4.85)}\\
\midrule
InternVL3-2B    & 40.18\\
+SFT            & 45.15 (+4.97)\\
+SFT, +GRPO           & 47.97 (+7.79)\\
+SFT, +HCM-GRPO       & \textbf{49.10 (+8.92)}\\
\bottomrule
\end{tabular}
\end{table}
\section{Conclusion}
In this work, we address the challenges of image screening with MLLMs by proposing a comprehensive solution that includes a novel dataset and an advanced training methodology. Our dataset evaluates physical plausibility reasoning ability across four critical dimensions, while our method leverages CoT data followed by HCM-GRPO reinforcement learning to significantly enhance performance. Notably, this task is highly challenging, as even leading closed-source MLLMs, such as GPT5.2 and Gemini3-Pro, perform poorly on the testing dataset. In contrast, our approach achieves superior results with a much smaller model, demonstrating the effectiveness of combining CoT data with reinforcement learning and the advantages of HCM-GRPO over the original GRPO. We believe these efforts will pave the way for more robust and reliable solutions in physical plausibility reasoning. Despite these achievements, some limitations still remain. \\
% \begin{itemize}
%\item 
Our model demonstrates good performance in recognizing appearance deformation and extension rationality, but it still struggles with placement layout as well as physical shadow. Appearance deformation and extension rationality cases can often be solved by detecting local, object-centric texture or shape cues and performing a relatively direct comparison with the reference image, making the decision largely dependent on fine-grained visual similarity. In contrast, physical shadow and placement layout require enforcing global physical constraints (e.g., coherent lighting direction, projection geometry, and stable support relations), for which isolated texture features of a single object are insufficient. For example, in the bad reasoning cases of Figure~\ref{Demonstration2}, the model correctly notices that candidates A and C of the left case contain shadows, yet fails to recognize that the shadows are inconsistent. Similarly, for the right case, the model can perceive that the bottom product touches the tabletop but cannot infer that a single-point contact combined with a tilted box violates physical stability. Future work will focus on enriching the CoT data with high-quality examples from different dimensions. This enhancement is expected to provide a stronger foundation for the reinforcement learning phase, further improving the model's reasoning capabilities in complex scenarios.\\
Since our dataset is centered on medicine packaging, the transferability of the learned capability may be limited when target objects differ substantially in geometry, appearance, or compositional complexity. This limitation points to the need for broader object diversity and cross-domain supervision. At the same time, the learned capability is not inherently restricted to this specific object category. The model primarily acquires more general physical reasoning patterns, including spatial configuration, relative position, occlusion, viewpoint consistency, and structural correspondence under partial observation. Therefore, insofar as other complex objects share these underlying spatial and physical dependencies, we expect the learned reasoning to generalize beyond the original domain. Accordingly, future work will focus on extending the benchmark to a wider range of object types, including those with irregular geometry, deformable structures, reflective materials, and more complex multi-object interactions. It will also be important to increase the diversity of scenes and environmental conditions, such as lighting variations, cluttered backgrounds, viewpoint changes, and severe occlusions, so that the benchmark can better reflect real-world scenarios and more rigorously evaluate the robustness of physical plausibility reasoning.\\
% \item Beyond generating higher-quality CoT data to target model weaknesses, we aim to achieve knowledge transfer in an unsupervised manner. In this paper, we do not extensively utilize the exploration split of our dataset. Consequently, a key direction for future work is to focus on enabling knowledge transfer for image screening using large volumes of unsupervised data. The ultimate goal is to allow existing MLLMs to automatically acquire domain-specific knowledge.
%\end{itemize}

\section{Declaration}
We present some declarations here. Regarding funding sources, this research does not receive any specific grant from funding agencies in the public, commercial, or not-for-profit sectors. During the preparation of this work, we use Qwen-VL-Max to polish the paper and correct grammatical errors. After using this tool, we review and edit the content as needed and take full responsibility for the content of the published article.

\newpage
\appendix

\section{More Benchmark Examples and Reasoning Cases}\label{apd:first}
\begin{figure}[h]
\centering
\includegraphics[width=0.85\textwidth]{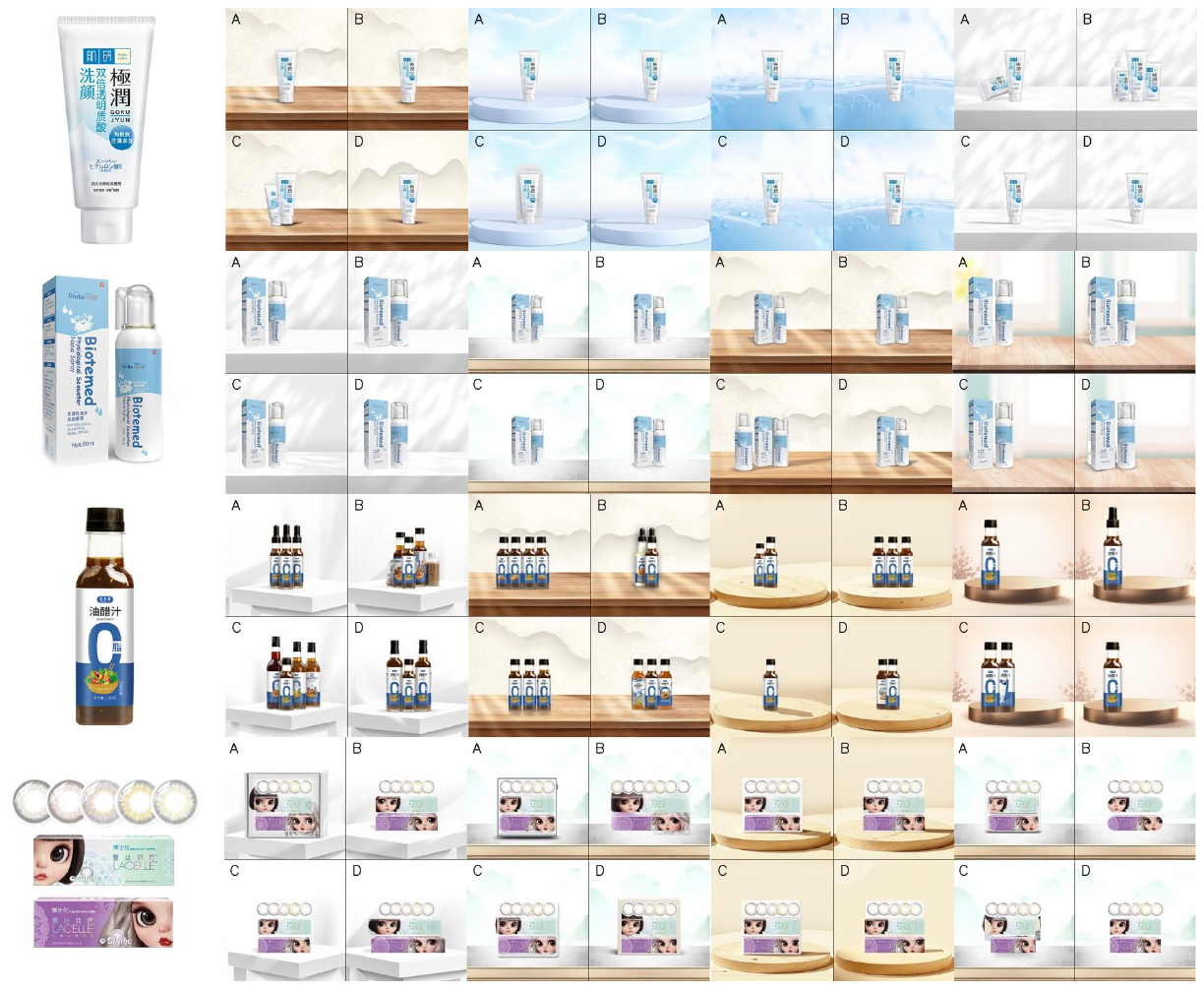}
\caption{Samples from the constructed dataset.} \label{Demonstration1}
\end{figure}

\begin{figure}[h]
\centering
\includegraphics[width=0.98\textwidth]{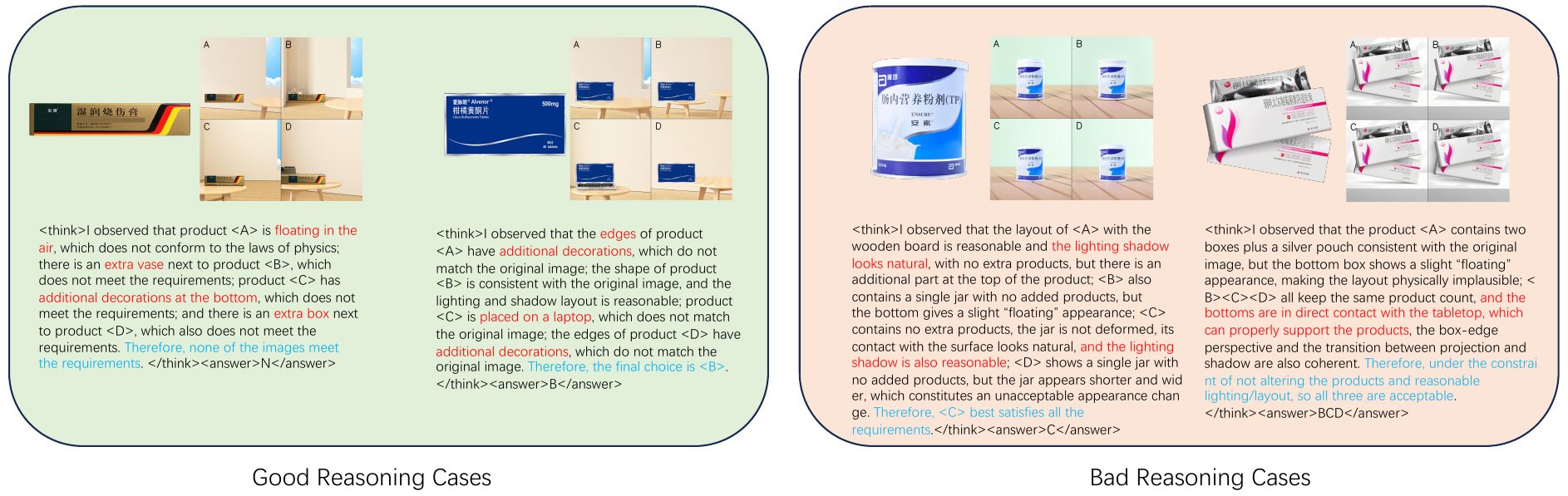}
\caption{Presentation of the reasoning cases.} \label{Demonstration2}
\end{figure}

We present some samples from the dataset here. Figure~\ref{Demonstration1} shows some original images along with their corresponding generated images. In Figure~\ref{Demonstration2}, we illustrate some reasoning cases of our model, with key words in the reasoning process highlighted in red and the decisions made by the model marked in blue.

\section{Prompt Templates}\label{apd:second}

We present all the prompts here, including weakly-supervised CoT prompt, answer-driven CoT prompt, and training and testing CoT prompts. There are two types of training and testing CoT prompts. One type directly asks the model to output an answer. The other type follows the GRPO thinking format, outputting the chains of thought process and the final answer. In all of our training and testing experiments, we use the following two types of prompts: CoT Format (highlighted in red) and Direct Format (highlighted in blue).

\begin{figure}[h]
\centering
\includegraphics[width=0.98\textwidth]{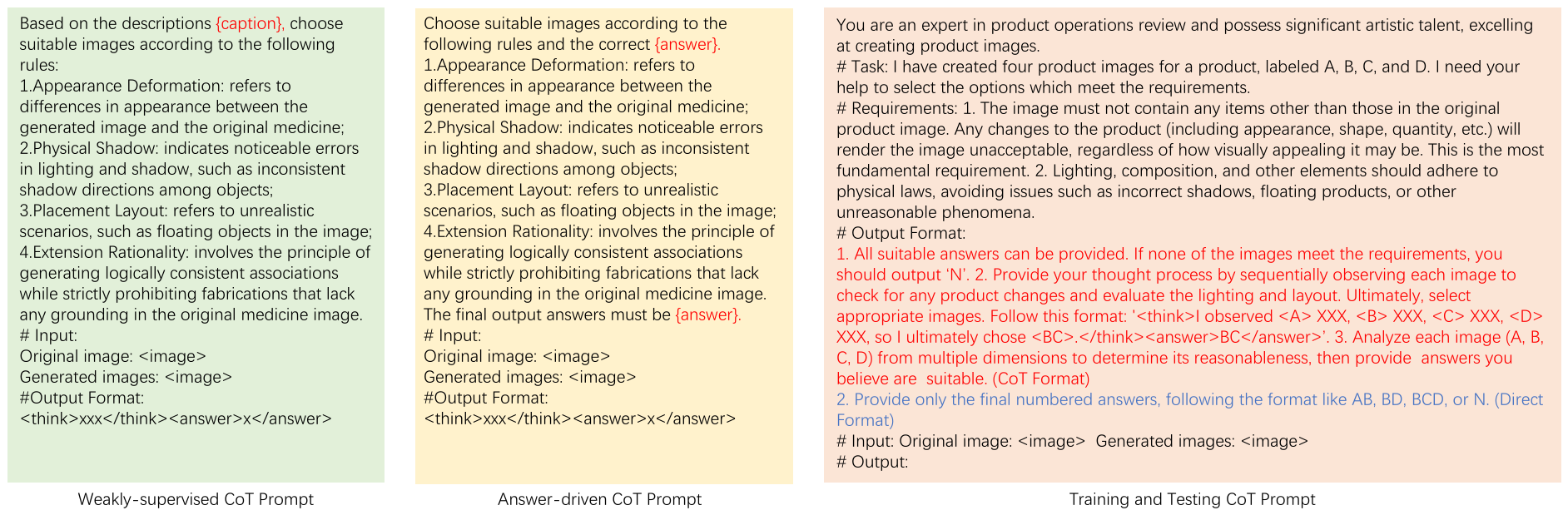}
\caption{Presentation of all the prompts.} \label{Demonstration3}
\end{figure}

\newpage
\bibliographystyle{elsarticle-num} 
\bibliography{ref}

\end{document}